\documentclass[10pt,twocolumn,letterpaper]{article}

\usepackage{cvpr}
\usepackage{times}
\usepackage{epsfig}
\usepackage{graphicx}
\usepackage{amsmath}
\usepackage{amssymb}
\usepackage{subfigure}
\usepackage{verbatimbox}
\usepackage{booktabs}

\usepackage[breaklinks=true,bookmarks=false]{hyperref}

\cvprfinalcopy 

\def\cvprPaperID{****} 

\begin{document}

\title{Diabetic Retinopathy Detection via Deep Convolutional Networks for Discriminative Localization and Visual Explanation}

\author{Zhiguang Wang, Jianbo Yang \thanks{The authors contribute equally to this work.}\\
GE Global Research\\
San Ramon, CA\\
{\tt\small \{zhiguang.wang, jianbo.yang\}@ge.com}
}

\maketitle

\begin{abstract}
We proposed a deep learning method for interpretable diabetic retinopathy (DR) detection. The visual-interpretable feature of the proposed method is achieved by adding the regression activation map (RAM) after the global averaging pooling layer of the convolutional networks (CNN). With RAM, the proposed model can localize the discriminative regions of an retina image to show the specific region of interest in terms of its severity level. We believe this advantage of the proposed deep learning model is highly desired for DR detection because in practice, users are not only interested with high prediction performance, but also keen to understand the insights of DR detection and why the adopted learning model works. In the experiments conducted on a large scale of retina image dataset, we show that the proposed CNN model can achieve high performance on DR detection compared with the state-of-the-art while achieving the merits of providing the RAM to highlight the salient regions of the input image.  
\end{abstract}

\section{Introduction}
Diabetes is an widespread disease in the world, and up to 2014 around 422 million people worldwide have this disease\footnote{http://www.who.int/mediacentre/factsheets/fs312/en/}. Diabetic retinopathy (DR) is an eye disease caused by the long-standing diabetes. Basically, DR affects blood vessels in the light-sensitive tissue (\emph{i.e.} retina). It becomes the leading cause of vision impairment and blindness for working-age adults in the world today\cite{aaaiss10Silberman}, and around half of Americans with diabetes have this disease to some extent. A widely-known challenge for DR is that it has no early warning sign, even for diabetic macular edema. Thus, it is highly desired that DR can be detected in time. Unfortunately, in practice the current DR detection solution is nearly infeasible to meet this requirement. Specifically, the current solution requires a well-trained clinician to manually evaluate digital color fundus photographs of retina, and DR is identified by locating the lesions associated with vascular abnormalities due to diabetes. Though this current solution is effective, it is  time-consuming and highly relies on the expertise of well-training practitioners. To solve this issue, in the past few years considerable efforts have been put on developing an automated solution for DR detection.

Most of previous automated solutions consists of two parts: feature extraction and detection/prediction algorithm \cite{TMI98Pinz,aaaiss10Silberman,Sensors09Akara,TBE06Wu}. Feature extraction is the main focus as standard machine learning algorithms can be directly used as the detection/prediction algorithm.  This type of approaches are effective to some extent  but also suffer from several shortcomings. First, as reviewed in Section \ref{sec:review}, the extracted features are all hand-crafted features. Thus, these features highly depend on the parameters of the used feature extraction tools and they are sensitive to the quality of fundus photography, like object view, exposedness, artifacts, noise, out-of-focus, etc. Second, feature extraction is a solo task rather than embedded into the whole DR detection framework. The above-mentioned features extraction methods can be considered as the universal image feature extraction methods that are applicable to most computer vision tasks, and they are not dedicated to the specific task, e.g., DR detection task considered in this paper.  It is worth noting that color fundus photography is more challenging than the standard scene or object images that most image feature extraction methods were developed based on, since the key signals are often tiny in fundus photography and they often look indiscriminating from noise and artifacts. Thus, these two challenges make it highly desirable to develop a systematical feature representation approach to effectively characterize the nature of features particularly related to the DR detection task.


Recently, the convolutional neural networks (CNN) has achieved tremendous success in computer vision area. It can model high-level abstractions in data relative to specific prediction task \cite{ Bengio2009FTML,Deng2014TSIP,lecun-89e,lecun-98b,Yang2015IJCAI}. In CNN, a multiple layers network is built up for automating feature design.  Specifically, each layer in deep architecture  performs a non-linear transformation on the outputs of the previous layer, so that the data are represented by a hierarchy of features from low-level to high-level. The key attribute of the CNN is conducting different processing units (\emph{e.g.} convolution, pooling, sigmoid/hyperbolic tangent squashing, rectifier and normalization ) alternatively. Such a variety of processing units can yield an effective nonlinear representation of local salience of the signals. Then, the deep architecture allows  multiple layers of these processing units to be stacked, so that this deep learning model can characterize the salience of signals in different scales. Also, in CNN, feature extraction and prediction algorithm are unified as a single model. Thus, the extracted features own more discriminative power, since the entire CNN model is trained under the supervision of output labels. Briefly speaking, the features extracted by the CNN are task dependent and non-handcrafted.

In this paper, we also adopt CNN as the key predictive algorithm, but aim to develop a more efficient CNN architecture that is particularly useful for large-scale dataset. Specifically, the CNN we built has no fully connected layer and only have convolutional and pooling layers. This setting significantly reduces the number of parameters (fully-connected layers often bring  more parameters than convolutional layers in the conventional CNN) and provides better conditions for interpretability of neural network as presented below. We show in experiments that with less parameters and no fully-connected layers the proposed CNN architecture can achieve the comparative prediction performance.  The key advantage of the proposed network structure is that it can provide a regression activation maps (RAM) of input image to show the contribution score of each pixel of input image for DR detection task. This RAM output, to some extent, somehow mitigates the well-known uninterpretable shortcoming of CNN as a black box method. We believe that this RAM output make the proposed solution more self-explained and can motivate the practitioners to trace the cause of the disease for every patient.

\section{Related Work}\label{sec:review}
The two-step (i.e., feature extraction and prediction) automated DR detection approaches dominated the field of DR detection for many years. Given color fundus photography, this type of approaches often extracted visual features from the images on the parts of blood vessels, fovea and optic disc \cite{TMI98Pinz,TBE06Wu}. The generic feature extraction methods developed in computer vision area were widely used here, e.g, hough transform, gabor filters and intensity variations. With the extracted features, an object detection or object registration algorithm like support vector machines and \emph{k}-NN were used to identify and localize exudates and hemorrhages \cite{aaaiss10Silberman,Sensors09Akara}. As mentioned before, this type of approaches are not as effective as the recent deep learning approaches, such as \cite{Antony15,Lim2014MAIHA,Pratta2016MIUA,Wang2015Neurocomputing}. All these deep learning approaches adopted the standard architecture like AlexNet and GoogLeNet to build their CNN, and based on the experimental results these deep learning approaches significantly outperform the traditional two-step approaches. Moreover, the recent DR detection competition held in Kaggle \cite{Kaggle15}\footnote{https://www.kaggle.com/c/diabetic-retinopathy-detection} witnessed that all top solutions adopted CNN as the key algorithm. However, all these CNN approaches require complex neural network structures, and it is hard for practitioners to understand the insight of CNN and clearly explain that which region of  the  color fundus photography is the main cause of the disease.

Understanding the insights of CNN has always been a pain point, though CNN yields excellent predictive performance. It is well-known that deriving theoretical results is quite challenging due to the nonliner and non-convex nature of CNN. To mitigate this issue, considerable efforts have been put on visualizing the CNN. A deconvolutional networks approach was proposed to visualize activated pattern in each hidden unit \cite{Zeiler13ECCV}. This method is limited as it is hard to summarize all hidden units's patterns into one pattern, and also only the hidden neurons in the hidden layers are analyzed though the networks considered also contain the fully-connected layers. The work \cite{Bazzani16WACV,Oquab14CVPR,Zhou15ICLR} and the reference therein include the objection location task besides the conventional object classification problem, so their CNN can predict the label of an image and also identify the region of the object related to the class label. Though this type of CNNs can predict the location of the object of interest, it still cannot reveal the insight of CNN. Recently, \cite{DB16CVPR,mahendran15CVPR} have presented the methods to invert the representation of images in each layer of the CNN. However, these approaches can only indicate what information is preserved in each layer of the CNN.

The most work most related to our method is \cite{Zhou16CVPR} in which class activation map is proposed to characterize the weighted activation maps after global average pooling or global maximum pooling layer. This idea has recently been generalized to time series analysis to localize the significant regions in the raw data \cite{wang2016time}. In this paper, we extend the method \cite{Zhou16CVPR} from a classification to a regression setting and shed light on DR detection problem.

\begin{table*}[!htbp]
	\centering
	\caption{Two proposed convolutional network structures: Net-5 and Net-4.}
	\begin{tabular}{rlr|rrr|rrr}
		&       &       & & Net-5 & & & Net-4 & \\
		\toprule
		& layer & Unit & Filter & Stride & Size & Filter & Stride & Size \\
		\midrule
		1     & Input   &       &       &       & 448      &       &       & 448  \\
		2     & Conv    & 32    & 5     & 2     & 224   & 4     & 2     & 224 \\
		3     & Conv    & 32    & 3     &       & 224   & 4     &       & 225 \\
		4     & MaxPool &       & 3     & 2     & 111   & 3     & 2     & 112 \\
		5     & Conv    & 64    & 5     & 2     & 56   & 4     & 2     & 56 \\
		6     & Conv    & 64    & 3     &       & 56    & 4     &       & 57 \\
		7     & Conv    & 64    & 3     &       & 56    & 4     &       & 56 \\
		8     & MaxPool &       & 3     & 2     & 27    & 3     & 2     & 27 \\
		9     & Conv    & 128   & 3     &       & 27    & 4     &       & 28 \\
		10    & Conv    & 128   & 3     &       & 27    & 4     &       & 27 \\
		11    & Conv    & 128   & 3     &       & 27    & 4     &       & 28 \\
		12    & MaxPool &       & 3     & 2     & 13    & 3     & 2     & 13 \\
		13    & Conv    & 256   & 3     &       & 13    & 4     &       & 14 \\
		14    & Conv    & 256   & 3     &       & 13    & 4     &       & 13 \\
		15    & Conv    & 256   & 3     &       & 13    & 4     &       & 14 \\
		16    & MaxPool &       & 3     & 2     & 6     & 3     & 2     & 6 \\
		17    & Conv    & 512   & 3     &       & 6     & 4     &       & 6 \\
		18    & Conv    & 512   & 3     &       & 6     & \multicolumn{1}{l}{N/a} &       & \multicolumn{1}{l}{N/a} \\
		19    & GlobalPool &       & \multicolumn{6}{c}{}  \\
		20    & Dense & 1     & \multicolumn{6}{c}{} \\
	\end{tabular}%
	\label{tab:arch}%
\end{table*}%

\section{Regression Activation Maps (RAM)}
Inspired by \cite{Zhou16CVPR}, we present in this section the idea of generating the RAM of an input image to localize the discriminative region towards the regression outcomes. It is known that the convolutional units of each layers of CNN act as visual concept detectors to identify low-level concepts like textures or materials,
to high-level concepts like objects or scenes. Deeper
into the network, the units become increasingly discriminative.
However, the fully-connected layers will make it difficult to identify the importance of different units for identifying the output labels (regression values, in our networks). Instead, using GAP and the linear output unit, we can directly visualize the  region of interest (ROI) that are most discriminative for a given regression value. As we use regression for the purpose of classification, each single RAM obtained for each single image explicitly depict the ROI on different clinical level.

The network architectures of our convolutional nets are shown in Table \ref{tab:arch}. Since we consider regression problem the output layer has one neuron outputting a real value. The key difference between our neural network and conventional neural networks like AlexNet \cite{Alex2012NIPS} and GoogLeNet \cite{Szegedy2015CVPR} lie in that our network uses global averaging pooling (GAP) layer to connect the last convolutional layer and the output layer, instead of using fully-connected layers. The idea of GAP layer is that each neuron in GAP obtains the spatial  average  of  the  feature  maps  from  the  last  convolutional layer so that the value of each neuron in GAP reflects its contribution to the final prediction. Specifically, supposing the last convolutional layer contains $K$ feature maps $\{g_k(i,j)|\forall i,j\},  k=1,...,K$ and $(i,j)$ is the spatial coordinate locating an entry in the feature map $k$. In the GAP layer, each feature map $g_k(i,j)$ in the last convolutional layer is mapped into a scaler $t_k$ by the function $t_k = \sum_{i,j}g_k(i,j)$. Then, the weighted sum of the output of the GAP layer $\hat{y} = \sum_{k=1}^Kt_kw_k$ is the value of the neuron in the output layer, where $\hat{y}$ is the predicted label and $w_k$ is the weight of neuron $k$ for the output of the global averaging pooling layer.

\begin{figure*}[!htbp]
\begin{center}
   \includegraphics[width=0.8\linewidth]{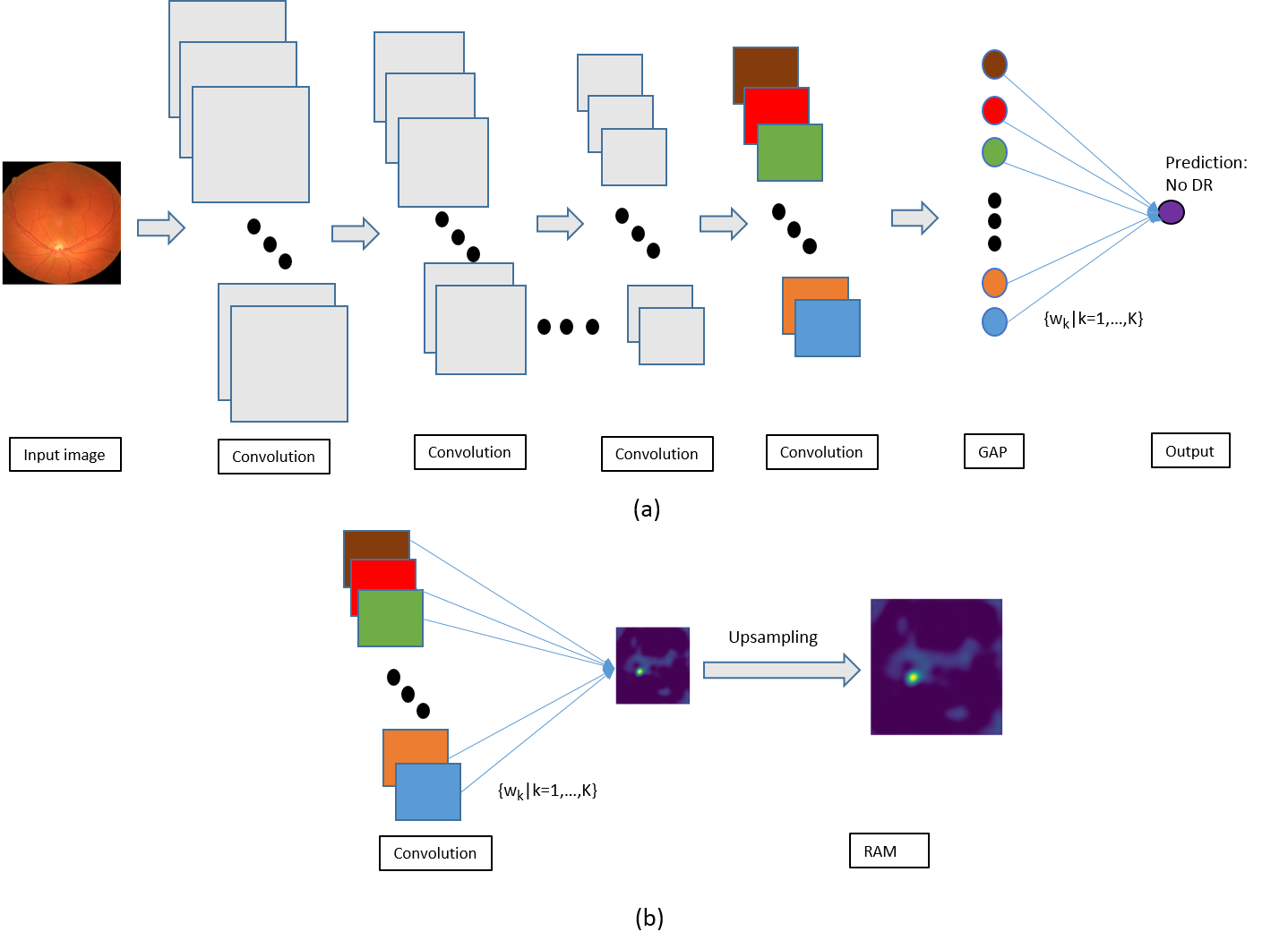}
\end{center}
   \caption{Illustration of adopted neural network structure (a) and regression activation mapping (b).}
   \label{fig:RAM}
\end{figure*}

Given the network structure in Table \ref{tab:arch}, the regression activation maps (RAM) is defined as below:
\begin{align*}
G(i,j) = \sum_{k=1}^Kg_k(i,j)w_k
\end{align*}
Thus, RAM is essentially a weighted sum of the feature maps in the last convolutional layer. The weights herein are the connections between the outputs of the global averaging pooling layer and the neuron in the output layer. Therefore, the final prediction can also expressed as:
\begin{align*}
\hat{y} = \sum_{k=1}^Kw_k\sum_{i,j}g_k(i,j) =\sum_{i,j} G(i,j)
\end{align*}
Intuitively, RAM contains the immediate information for prediction (feature maps in the last convolutional layer and weights before the final regression output), and also maintain the correspondence between last convolutional feature map and input images. Therefore, RAM can localize the discriminative region towards the regression outcomes. The illustration of the adopted neural network structure and RAM are show in Figure \ref{fig:RAM}.

\section{Experiments}
\begin{figure*}[!htbp]
    \centering
    \begin{subfigure}
        \centering
        \includegraphics[height=1in]{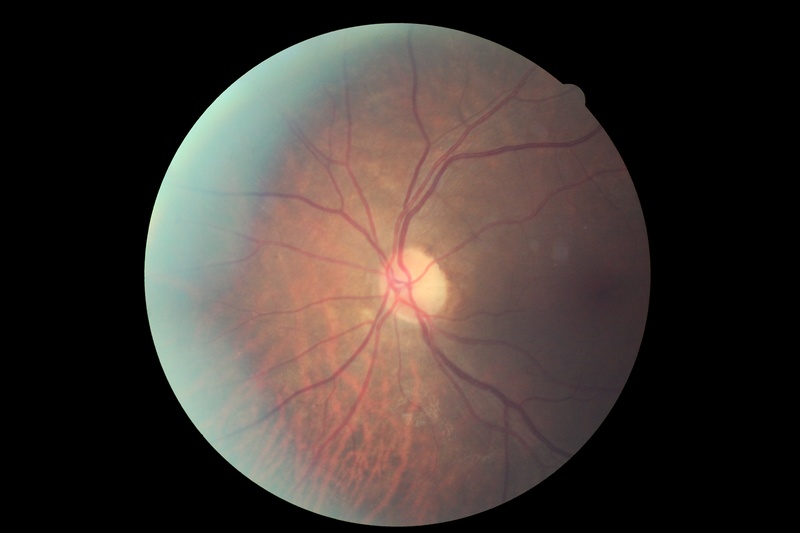}
    \end{subfigure}%
    \begin{subfigure}
        \centering
        \includegraphics[height=1in]{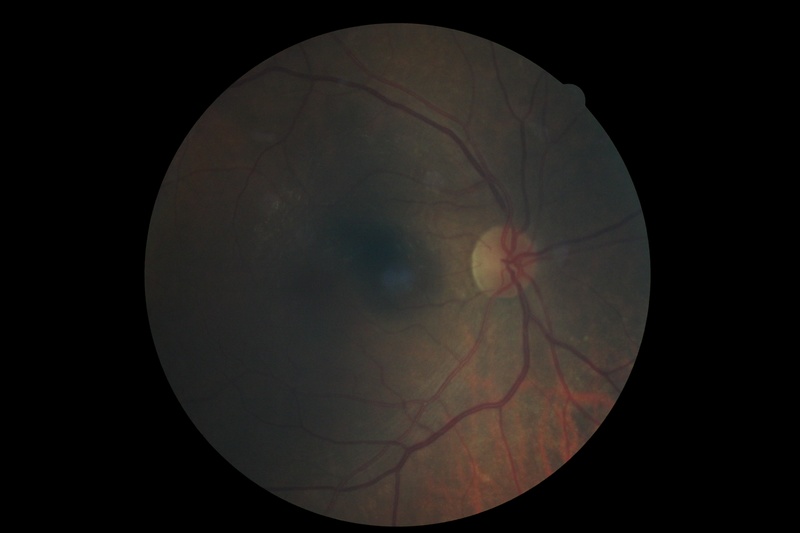}
    \end{subfigure}
    \begin{subfigure}
        \centering
        \includegraphics[height=1in]{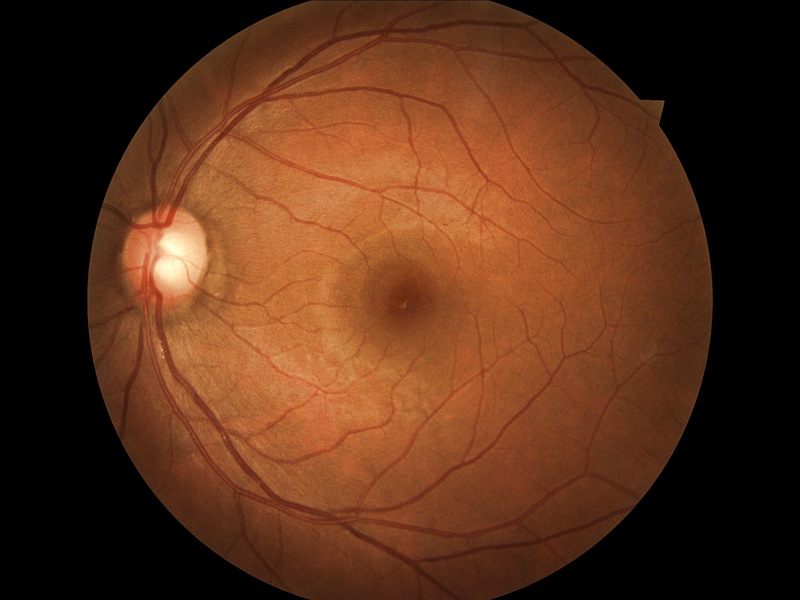}
    \end{subfigure}%
    \begin{subfigure}
        \centering
        \includegraphics[height=1in]{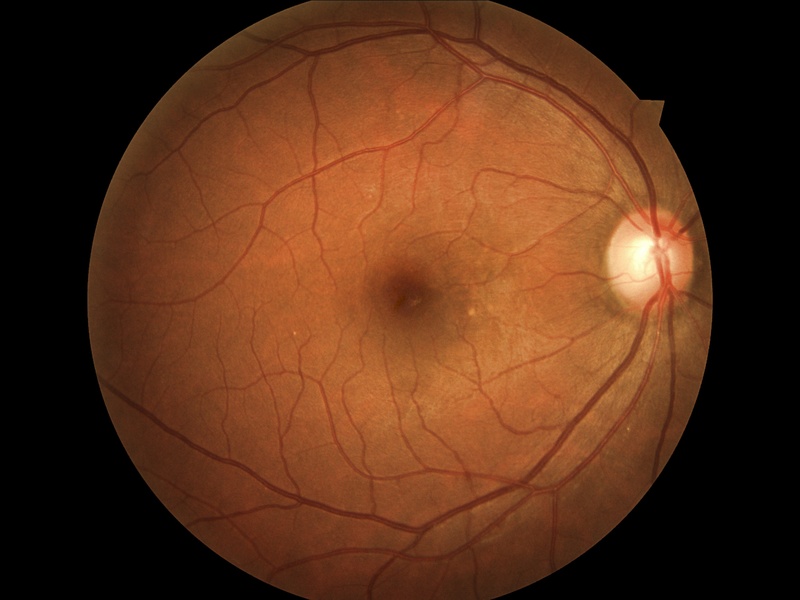}
    \end{subfigure}
    \caption{Sample images of color retina images dataset.}
    \label{fig:samples}
\end{figure*}

\subsection{Datasets}
\begin{figure}[t]
\begin{center}
   \includegraphics[width=1\linewidth]{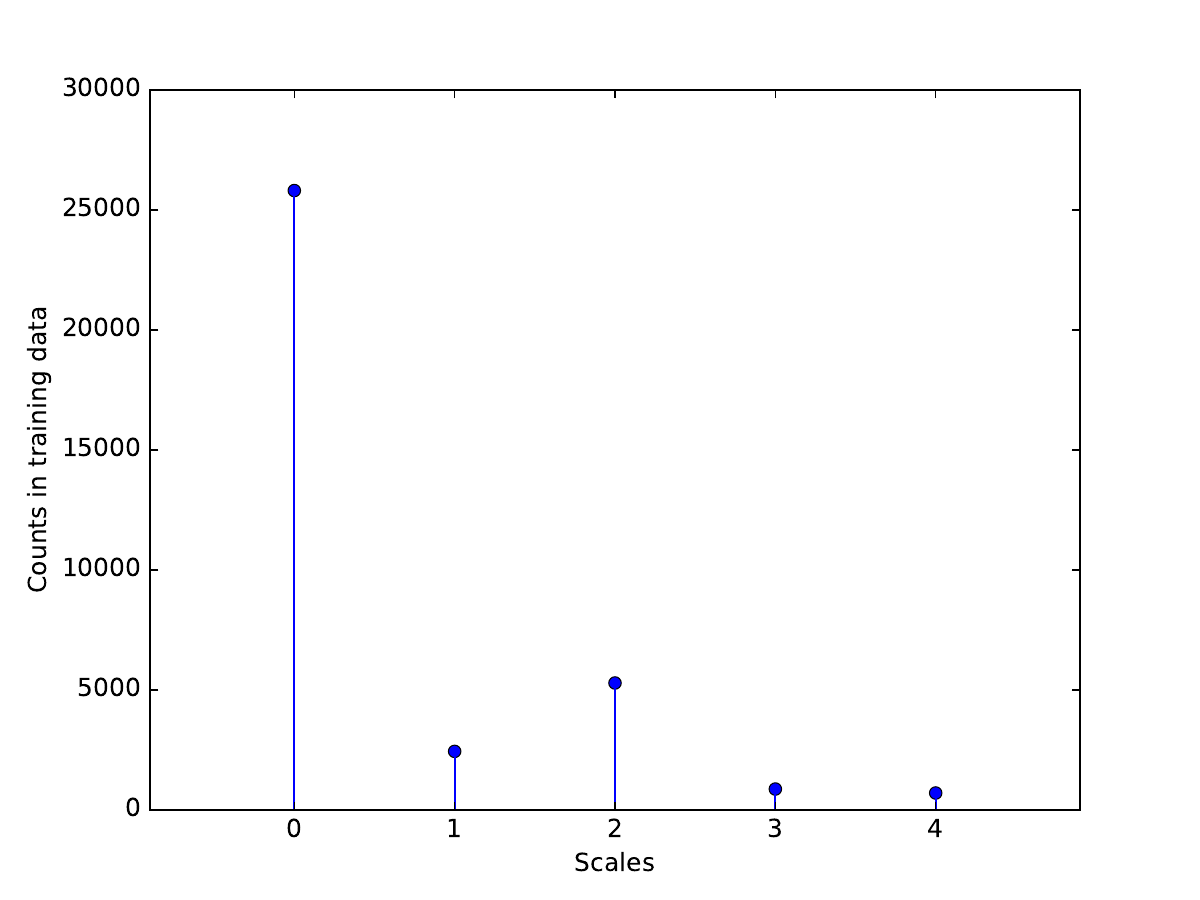}
\end{center}
   \caption{Counts of images for different scales in the training dataset.}
\label{fig:counts}
\end{figure}

The color retina images are downloaded from the Kaggle website\footnote{https://www.kaggle.com/c/diabetic-retinopathy-detection/data}. The training dataset contains 35126 high resolution images under a variety of imaging conditions. These retina images were obtained from a group of subjects, and for each subject two images were obtained for left and right eyes, respectively. The labels were provided by clinicians who rated the presence of diabetic retinopathy in each image by a scale of ``0, 1, 2, 3, 4", which represent ``no DR", ``mild", ``moderate", ``severe", ``proliferative DR" respectively. As mentioned in the description of the dataset, the images in the dataset come from different models and types of camera, which can affect the visual appearance of left vs. right. The samples images are shown in Fig \ref{fig:samples}. Also, the dataset doesn't have the equal distributions among the 5 scales. As one can expect, normal data with label ``0" is the biggest class in the whole dataset, while ``proliferative DR" data is the smallest class.  Fig \ref{fig:counts} shows counts of images for different scales in the training dataset.

\begin{table*}[!htbp]
	\centering
	\caption{Performance statistics of the benchmark and our approaches on test dataset.}
	\begin{tabular}{l|r|r}
		& Baseline & Ours \\
		\toprule
		Kappa score (Public Leaderboard) & 0.8542 & 0.85034 \\
		Kappa score (Private Leaderboard) & 0.8448 & 0.8412 \\
		\midrule
		Parameter \# (net-5) & 12.4M & 9.7M \\
		Training time (second/epoch) & 422.1 & 367.3 \\
		\midrule
		Parameter \# (net-4) & 12.5M & 9.8M \\
		Training time (second/epoch) & 451.7 & 398.2 \\
		\midrule
		RAM & No & Yes \\
	\end{tabular}%
	\label{tab:stats}%
\end{table*}%

\subsection{Benchmark method}
\label{sec:baseline}
We consider the publicly-available method \cite{Antony15} as the benchmark method, which was ranked as the second place in the Kaggle competition. This method crops away all background and resize the images to squares of 128, 256 and 512 pixels.
The interested readers may refer to \cite{Antony15} for more detailed settings of the baseline method. We summarize the main features of the baseline methods as follow:
\begin{description}
  \item[Resampling] First, sample all classes such that all classes are represented equally on average. Then,  gradually reduce the over sampling of rare classes.
  \item[Initialization and pretraining] Orthogonal initialization is used to initialize weights and biases. First, train smaller networks on 128 pixel images. Then, use the trained weights to (partially) initialize medium networks for training on 256 pixel images. Finally, use the trained weights of medium networks to (partially) initialize large network for training on 512 pixel images.
  \item[Data augmentation] The common image transformation like translation, stretching, rotation, flipping and color augmentation are used for data augmentation.
  \item[Feature blending] Use the last pooling layer of the convolutional networks as extracted features, and then blend all these extracted features from 50 networks outputs (with different augmentation). Then, a fully-connected neural network is used on the blended features as the final predictor.
\end{description}

\subsection{Experimental Settings}
We trained our convolutional neural network in Table \ref{tab:arch} on a single Tesla-P100 GPU. For nonlinearity, we use leaky (0.01) rectifier units following each convolutional layer. The networks are trained with Nesterov momentum with fixed schedule over 250 epochs. For the nets on 256 and 128 pixel images, we stop training after 200 epochs. L2 weight decay with factor 0.0005 are applied to all layers. As we treat the problem as a regression problem, the loss function is mean squared error.The convolutional networks have untied biases. Batch size is fixed at 32 for all networks. \footnote{Codes are available at https://github.com/cauchyturing/kaggle\_diabetic\_RAM.}. Following the evaluation setting \cite{Kaggle15}, the quadratic weighted Kappa score is adopted as the performance metric of prediction. Specifically, the predicted regression values are discretized at the thresholds $(0.5, 1.5, 2.5, 3.5)$ to obtain integer levels for computing the Kappa scores and making submissions.  All the features mentioned in Section \ref{sec:baseline} were also adopted in our model training.

\begin{figure*}[t]
	\begin{center}
		\includegraphics[width=0.9\linewidth]{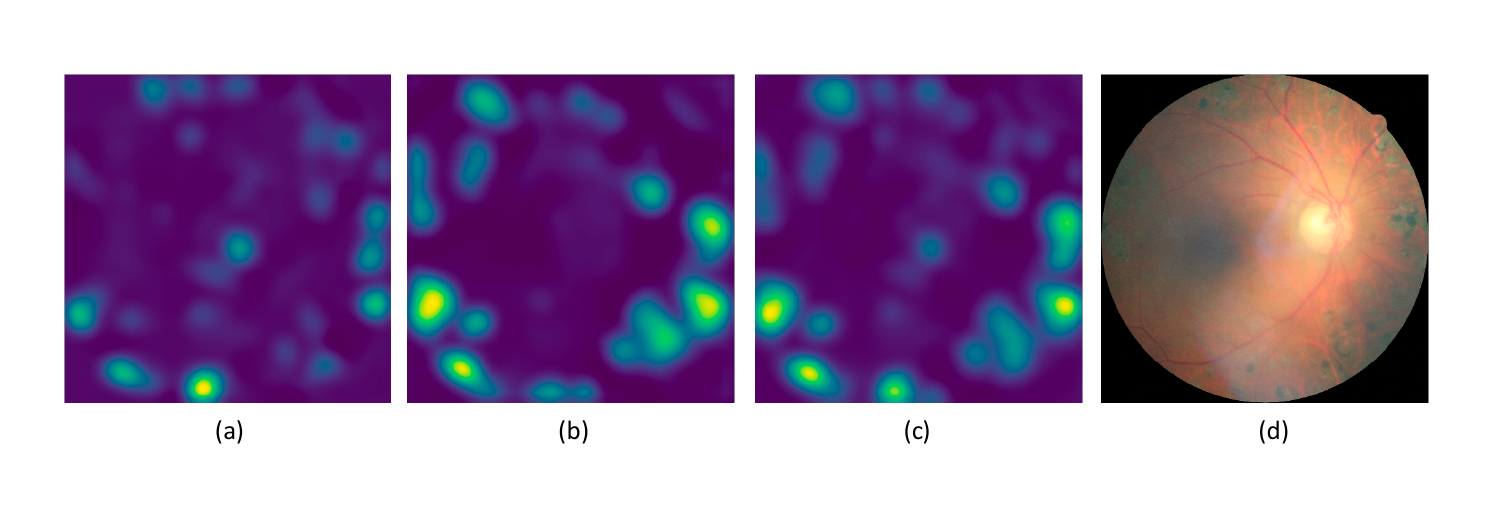}
	\end{center}
	\caption{The example RAM generated from the (a). 128 pixel image and (b). 256 pixel image. Note that the RAM output by the neural network is of size 56X56, and they are the up-sampled by Lanczos interpolation functions as shown in plots (a) and (b). The ensembled RAM averaged from both resolutions are shown in (c).(d) is the original image.}
	\label{fig:syncram}
\end{figure*}

\subsection{Performance of Kappa Score}
Following \cite{Antony15}, we split 35126 images into training and validation datasets in a ratio of 9 to 1 for local evaluation purpose, and we also submit our prediction results on the test dataset to Kaggle to obtain the Kappa score. Table \ref{tab:stats} summarizes the performance of both benchmark and our approach on the test dataset.  By simply replacing the fully-connected layer with the global average pooling layer, our networks achieved very competitive Kappa score compared with the benchmark while reducing the parameter size by about 21.8\% and speed-up the training by 11.8\%-13.1\%. As mentioned in Section \ref{sec:baseline}, feature blending strategy was used. Thus, the final Kappa score is an average of six per patient blends for the two convolutional network architectures and three different sets of trained weights.

Considering that the key signals in the retina images like Micro-aneurysms are very small with respect to the retina, we also evaluate the performance of the our model with respect to different size of input images.  On the validation set, our networks achieve the Kappa scores of around 0.70 for 256 pixel images,  0.80 for 512 pixel images and 0.81 for 768 pixel images on both Net-5 and Net-4 settings without feature blending. This observation suggests that the larger input images the better prediction performance but there is no much gain when the image size is greater than 512. Taking computation cost into consideration, we limit the input image size to be not greater than 512.

\begin{figure*}[!htbp]
	\begin{center}
		\includegraphics[width=1\linewidth]{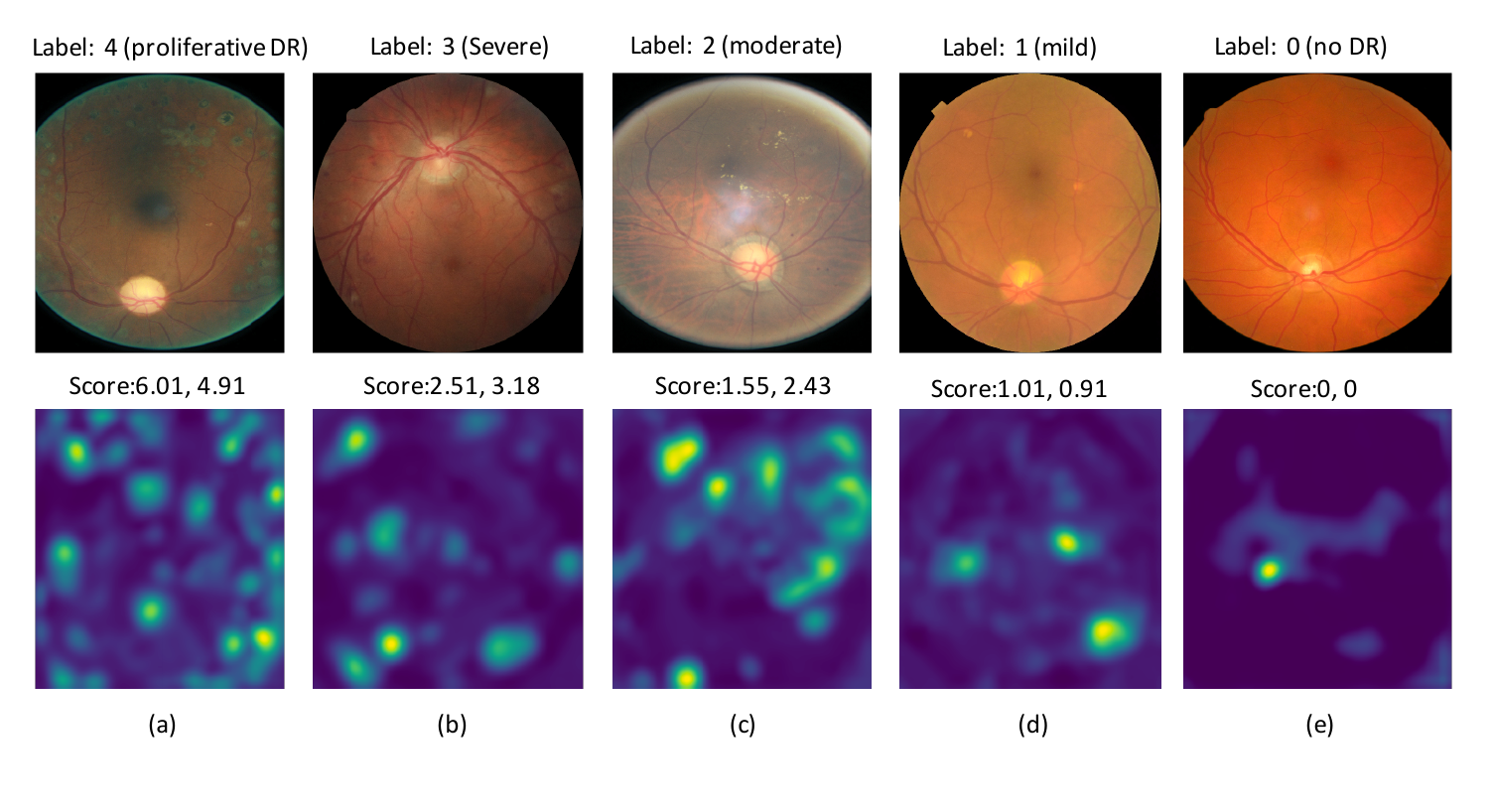}
	\end{center}
	\caption{Ground truth and the corresponding RAMs. The two scores are from the 128 and 256 pixel images, respectively.}
	\label{fig:levelRAM}
\end{figure*}

\begin{figure*}[!htbp]
	\begin{center}
		\includegraphics[width=0.9\linewidth]{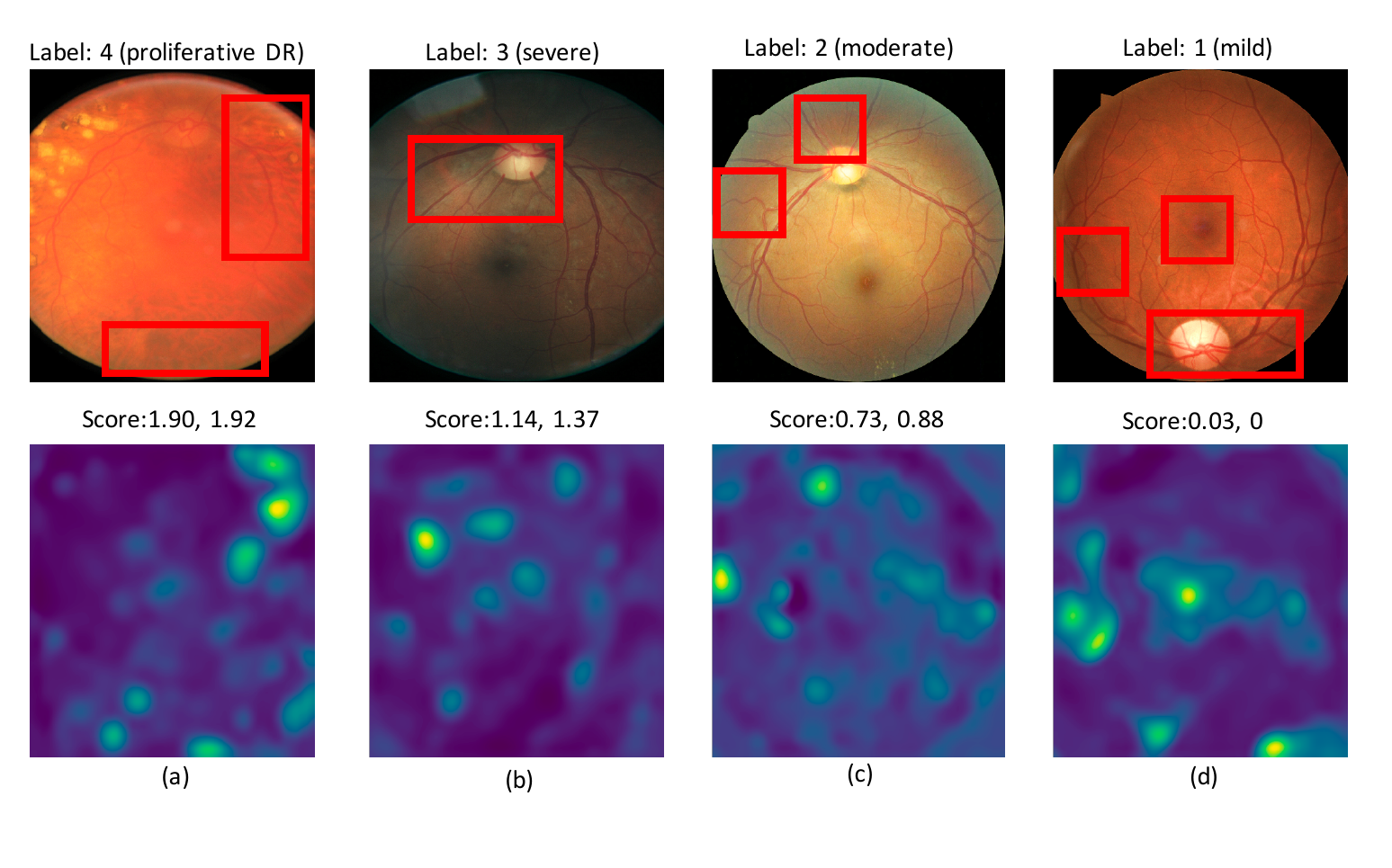}
	\end{center}
	\caption{Examples of the conservative diagnostic by the neural networks on different severity levels.}
	\label{fig:ConserveRAM}
\end{figure*}

\subsection{Discriminative Localization by RAM}

\subsubsection{Network settings for RAM}

To generate RAM, we used Net-5 with the 128 and 256 pixel images as the input. We also removed several convolutional layers of Net-5 for each input size to increase the resolution of RAM, since the localization ability of RAM can be significantly improved
when the last convolutional layer before GAP had a
higher spatial resolution \cite{Zhou16CVPR}. Specifically, we made the following modifications: For Net-5 on the 128 pixel images, we removed the layers after Conv-11 and all the strides excepts the Maxpool-8, which resulting in a mapping resolution of $54 \times 54$. For Net-5 on the 256 pixel images, we removed the layers after Conv-15 and last two max pooling layer, which resulted in a mapping resolution of $56 \times 56$. Each of these networks were then fine-tuned \footnote{Train from the scratch has the similar performance but a little bit slower.} on the training data.

\subsubsection{Fusion of multiple RAMs }

Figure \ref{fig:syncram}(a) and (b) show the RAMs from the input images of size 128 and 256 pixels respectively and their corresponding Kappa scores 5.06 and 4.63. We noticed that these RAMs reflect different ROIs that may both have contributions to the final Kappa score prediction. Thus, we consider the fusion of multiple RAMs generated from the different scales of input images. The fusion of the RAMs (a) and (b), i.e., the average of the values in RAM matrices, is plotted in Figure \ref{fig:syncram}(c). By referring to the original image shown in Figure \ref{fig:syncram}(d), we argue that Figure \ref{fig:syncram}(c) can better capture the ROI of original image. We also found the similar phonomania from other examples. So we conclude that the fusion of different RAMs from various resolutions is simple and effective to depict the comprehensive ROI, so the fused RAM is only reported in the following analysis.

\subsubsection{Analysis on RAM}

For the mild-conditioned patients, RAM learned to discover the narrowing of the retinal arteries associated with reduced retinal blood flow (Figure \ref{fig:levelRAM}(d)), where the vessel shows dark red. The dysfunction of the neurons of the inner retina, followed in later stages (moderate) by changes in the function of the outer retina are captured in Figure \ref{fig:levelRAM}(c), as such dysfunction protects the retina from many substances in the blood (including toxins and immune cells), leading to the leaking of blood constituents into the retinal neuropile. When the patients belong to the next stage (severe), as the basement membrane of the retinal blood vessels thickens, capillaries degenerate and lose cells leading to loss of blood flow and progressive ischemia and microscopic aneurysms which appear as balloon-like structures jutting out from the capillary walls. RAM, as shown in Figure \ref{fig:levelRAM}(b), learned to converge its focus on the border where the balloon-like structures occurs. As the disease progresses to the proliferative stage, the lack of oxygen in the retina causes fragile, new, blood vessels to grow along the retina and in the clear, gel-like vitreous humour that fills the inside of the eye. In Figure \ref{fig:levelRAM}(a), RAM
shows the model put its attention on the grey dots scattering around, which undoubtedly demonstrate the proliferative stage. We also note that if the patient has no DR and the score predicted by the model is smaller than 0.5, then the RAM uniformly shows the dot-like focus near the pupil (\ref{fig:levelRAM}(e)).

We also observed that the proposed model may have the conservative diagnostics behavior, which means that the predicted regression value is often smaller than the ground truth. The examples of this case are plotted in Figure \ref{fig:ConserveRAM}. One plausible reason is that the training dataset has considerably large number of normal images as shown in Figure \ref{fig:counts} so the learned model might be biased to the ``0" class. Other reasons could be due to image quality and the sensitivity on the ambiguous features between different levels. Specifically, in Figure \ref{fig:ConserveRAM}(a), though our model ignores the proliferate on the top left corner, it can capture the leaking of blood in the region of retinal neuropile. To identify leaking of blood in retinal neuropile may even be challenging for the clinicians. In (b) and (c), the original images contain very limited information about the balloon-like jitter and the blood leaking due to partial exposedness, so the RAM are scattered on the dard red vessels which are caused by the reduced retinal blood flow.

Based on the above analysis, RAM provides the reasonable transparency on our deep learning model to see why and how it makes the decision. The visual explanation of RAM may assist the clinicians to quickly identify the pathogenesis of disease.

\section{Conclusions}
Practically, clinicians can identify DR by the presence of lesions associated with the vascular abnormalities caused by the disease. While this approach is effective, its resource demands are high. In this work, we provided a deep learning model that includes regression activation maps layer (RAM).  The RAM layer can provide the robust interpretability of the proposed detection model by monitoring the pathogenesis so that the proposed model can be taken as an assistant for clinicians. With this feature, the proposed model can still yield the competitive performance of DR detection, compared with the state-of-the-art methods. In future, we would consider to extend the proposed method to other medical application problems.

\bibliographystyle{ieee}
\bibliography{egbib}

\begin{thebibliography}{10}\itemsep=-1pt

\bibitem{Kaggle15}
{Diabetic Retinopathy Detection | Kaggle:
  https://www.kaggle.com/c/diabetic-retinopathy-detection}, 2015.

\bibitem{Antony15}
M.~Antony and S.~Brüggemann.
\newblock {Kaggle Diabetic Retinopathy Detection: Team o$\_$O solution}, 2015.

\bibitem{Bazzani16WACV}
L.~Bazzani, A.~Bergamo, D.~Anguelov, and L.~Torresani.
\newblock Self-taught object alization with deep networks.
\newblock In {\em 2016 IEEE Winter Conference on Applications of Computer
  Vision (WACV)}, pages 1--9, 2016.

\bibitem{Bengio2009FTML}
Y.~Bengio.
\newblock Learning deep architectures for {AI}.
\newblock {\em Found. Trends Mach. Learn.}, 2(1):1--127, Jan. 2009.

\bibitem{Deng2014TSIP}
L.~Deng.
\newblock A tutorial survey of architectures, algorithms, and applications for
  deep learning.
\newblock {\em APSIPA Transactions on Signal and Information Processing}, 2014.

\bibitem{DB16CVPR}
A.~Dosovitskiy and T.Brox.
\newblock Inverting visual representations with convolutional networks.
\newblock In {\em CVPR}, 2016.

\bibitem{Alex2012NIPS}
A.~Krizhevsky, I.~Sutskever, and G.~E. Hinton.
\newblock Imagenet classification with deep convolutional neural networks.
\newblock In {\em NIPS}, 2012.

\bibitem{lecun-89e}
Y.~LeCun, B.~Boser, J.~S. Denker, D.~Henderson, R.~E. Howard, W.~Hubbard, and
  L.~D. Jackel.
\newblock Backpropagation applied to handwritten zip code recognition.
\newblock {\em Neural Computation}, 1(4):541--551, Winter 1989.

\bibitem{lecun-98b}
Y.~LeCun, L.~Bottou, G.~Orr, and K.~Muller.
\newblock Efficient backprop.
\newblock {\em Neural Networks: Tricks of the trade}, pages 9 -- 50, 1998.

\bibitem{Lim2014MAIHA}
G.~Lim, M.~L. Lee, W.~Hsu, and T.~Y. Wong.
\newblock Transformed representations for convolutional neural networks in
  diabetic retinopathy screening.
\newblock In {\em AAAI Workshop on Modern Artificial Intelligence for Health
  Analytics (MAIHA), AAAI}, 2014.

\bibitem{mahendran15CVPR}
A.~Mahendran and A.~Vedaldi.
\newblock Understanding deep image representations by inverting them.
\newblock In {\em CVPR}, 2015.

\bibitem{Oquab14CVPR}
M.~Oquab, L.~Bottou, I.~Laptev, and J.~Sivic.
\newblock Learning and transferring mid-level image representations using
  convolutional neural networks.
\newblock In {\em CVPR}, 2014.

\bibitem{TMI98Pinz}
A.~Pinz, S.~Bernogger, P.~Datlinger, and A.~Kruger.
\newblock Mapping the human retina.
\newblock {\em IEEE Transactions on Medical Imaging}, 17(4):606--619, Aug 1998.

\bibitem{Pratta2016MIUA}
H.~Pratta, F.~Coenenb, D.~M. Broadbentc, S.~P. Hardinga, and Y.~Zheng.
\newblock Convolutional neural networks for diabetic retinopathy.
\newblock In {\em International Conference On Medical Imaging Understanding and
  Analysis ({MIUA})}, 2016.

\bibitem{aaaiss10Silberman}
N.~Silberman, K.~Ahrlich, R.~Fergus, and L.~Subramanian.
\newblock Case for automated detection of diabetic retinopathy.
\newblock In {\em AAAI Spring Symposium: Artificial Intelligence for
  Development}. AAAI, 2010.

\bibitem{Sensors09Akara}
A.~Sopharak, B.~Uyyanonvara, and S.~Barman.
\newblock Automatic exudate detection from non-dilated diabetic retinopathy
  retinal images using fuzzy c-means clustering.
\newblock {\em Sensors}, 9(3):2148--2161, 2009.

\bibitem{Szegedy2015CVPR}
C.~Szegedy, W.~Liu, Y.~Jia, P.~Sermanet, S.~Reed, D.~Anguelov, D.~Erhan,
  V.~Vanhoucke, and A.~Rabinovich.
\newblock Going deeper with convolutions.
\newblock In {\em CVPR}, 2015.

\bibitem{Wang2015Neurocomputing}
S.~Wang, Y.~Yin, G.~Cao, B.~Wei, Y.~Zheng, and G.~Yang.
\newblock Hierarchical retinal blood vessel segmentation based on feature and
  ensemble learning.
\newblock {\em Neurocomputing}, 149:708--717, 2015.

\bibitem{wang2016time}
Z.~Wang, W.~Yan, and T.~Oates.
\newblock Time series classification from scratch with deep neural networks: A
  strong baseline.
\newblock {\em arXiv preprint arXiv:1611.06455}, 2016.

\bibitem{TBE06Wu}
D.~Wu, M.~Zhang, J.-C. Liu, and W.~Bauman.
\newblock On the adaptive detection of blood vessels in retinal images.
\newblock {\em IEEE Transactions on Biomedical Engineering}, 53(2):341--343,
  Feb 2006.

\bibitem{Yang2015IJCAI}
J.~B. Yang, M.~N. Nguyen, P.~P. San, X.~L. Li, and S.~Krishnaswamy.
\newblock Deep convolutional neural networks on multichannel time series for
  human activity recognition.
\newblock In {\em IJCAI}, 2015.

\bibitem{Zeiler13ECCV}
M.~D. Zeiler and R.~Fergus.
\newblock Visualizing and understanding convolutional networks.
\newblock In {\em ECCV}, 2014.

\bibitem{Zhou15ICLR}
B.~Zhou, A.~Khosla, {\`{A}}.~Lapedriza, A.~Oliva, and A.~Torralba.
\newblock Object detectors emerge in deep scene cnns.
\newblock In {\em ICLR}, 2015.

\bibitem{Zhou16CVPR}
B.~Zhou, A.~Khosla, {\`{A}}.~Lapedriza, A.~Oliva, and A.~Torralba.
\newblock Learning deep features for discriminative localization.
\newblock In {\em CVPR}, 2016.

\end{thebibliography}

\end{document}